# Diffusion-SAFE: Diffusion-Native Human-to-Robot Driving Handover for Shared Autonomy

Yunxin Fan[1], Monroe Kennedy III[1,2]

***Abstract*—Shared autonomy in driving requires anticipating human behavior, flagging risk before it becomes unavoidable, and transferring control safely and smoothly. We propose DIFFUSION-SAFE, a closed-loop framework built on two diffusion models: an *evaluator* that predicts multimodal human-intent action sequences for probabilistic risk detection, and a *safety-guided copilot* that steers its denoising process toward safe regions using the gradient of a map-based safety certificate. When risk is detected, control is transferred through *partial diffusion*: the human plan is forward-noised to an intermediate level and denoised by the safety-guided copilot. The forward-diffusion ratio $\rho$ acts as a continuous takeover knob—small $\rho$ keeps the output close to human intent, while increasing $\rho$ shifts authority toward the copilot, avoiding the mixed-unsafe pitfall of action-level blending. Unlike methods relying on hand-crafted score functions, our diffusion formulation supports both safety evaluation and plan generation directly from demonstrations. We evaluate DIFFUSION-SAFE in simulation and on a real ROS-based race car, achieving 93.0%/87.0% (sim/real) handover success rates with smooth transitions.**

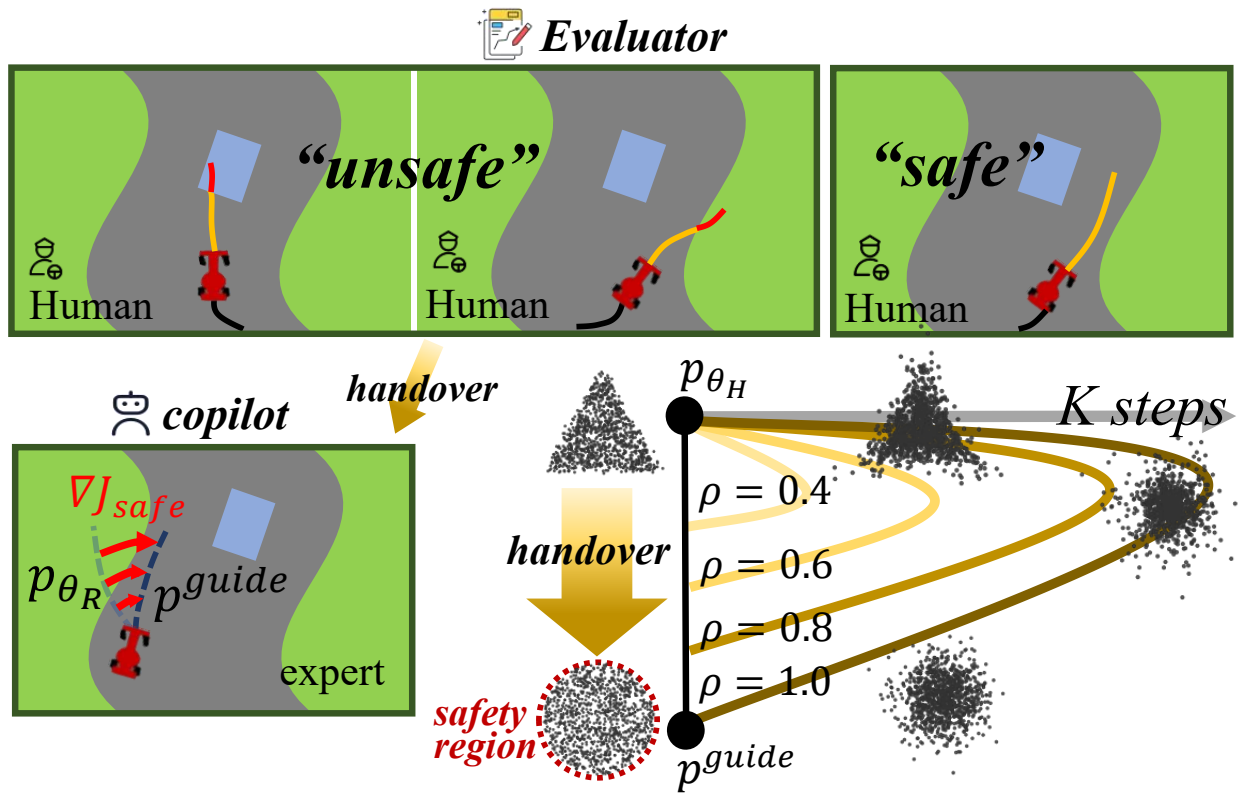

Fig. 1. Handover is realized via partial diffusion: the human plan is forward-noised and denoised by the safety-guided copilot, with the forward-diffusion ratio $\rho$ acting as a continuous takeover knob—small $\rho$ preserves human intent, while $\rho$=1 yields full copilot control. The evaluator predicts multimodal human-intent distributions for risk detection.

## I. INTRODUCTION

Shared autonomy bridges full autonomy and complete manual operation by enabling a human user and an assistive agent to collaboratively control a robot [1], [2]. In driving tasks, the assistive system must infer near-future human intent from interaction history and initiate a progressive handover before unsafe states become unavoidable. However, human behavior is inherently inconsistent, multimodal, and can change abruptly under out-of-distribution (OOD) conditions, making intent modeling and timely intervention challenging [3]. A further difficulty is that a smooth handover is not necessarily safe: many shared-control pipelines pursue smoothness via action-level blending [4], yet in driving the safe-action set is often nonconvex, so convex blending of two individually safe actions can produce an unsafe outcome [5]. For instance, averaging a left and a right evasive steer may drive straight into the obstacle. These observations call for a handover design that is simultaneously predictive, smooth, and safe.

*Handover in shared autonomy.* Current handover strategies rely on momentary driver-state cues (e.g., gaze or posture) [6] or hand-tuned thresholds that lack long-horizon predictive capability. Beyond the triggering question, the mechanism of control transfer itself poses challenges. At the control level, *hard-switching* methods transfer full authority once a monitoring threshold is crossed, but the resulting discontinuity can startle the driver and degrade ride comfort [7]. *Blending* methods interpolate between human and robot commands at the action level [1], [4], achieving smoother transitions but reintroducing the mixed-unsafe failure mode whenever the safe-action set is nonconvex [5]. An alternative to action-level interpolation operates in the plan space. SDEdit [8] offers a principled mechanism for blending a source signal with a learned prior via partial diffusion. Yoneda et al. [9] apply this idea to shared autonomy, using the corruption level to modulate the degree of autonomy without action-level mixing. However, their formulation provides no mechanism to enforce safety during the reverse process, nor a predictive trigger for initiating handover.

*Safety in diffusion-based planning.* Diffusion models have emerged as expressive trajectory generators in robotics and autonomous driving [10], [11], owing to their ability to capture multimodal distributions and to accept test-time guidance without retraining. However, ensuring safety during diffusion-based trajectory generation remains an open challenge. SafeDiffuser [12] enforces Control Barrier Function (CBF) constraints through a quadratic programming (QP) at each diffusion step, while Constrained Diffusers [13] formulates safety as a constrained Langevin sampling problem using projected, primal-dual, and augmented Lagrangian methods without retraining. These methods demonstrate that safety can be injected into the reverse process, yet they address a single autonomous planner and do not consider the shared-autonomy setting where control authority must be *progressively transferred* [7].

To address these gaps, we propose DIFFUSION-SAFE, a closed-loop shared autonomy framework built on two diffusion

This work was supported by Stanford CARS. The code will be released upon publication. The authors are members of the Departments of [1]Mechanical Engineering and [2]Computer Science, Stanford University, Stanford, CA, 94305. {yunxin6, monroek}@stanford.edu. Project website: https://codebase2026.github.io/Diffusion-SAFE/

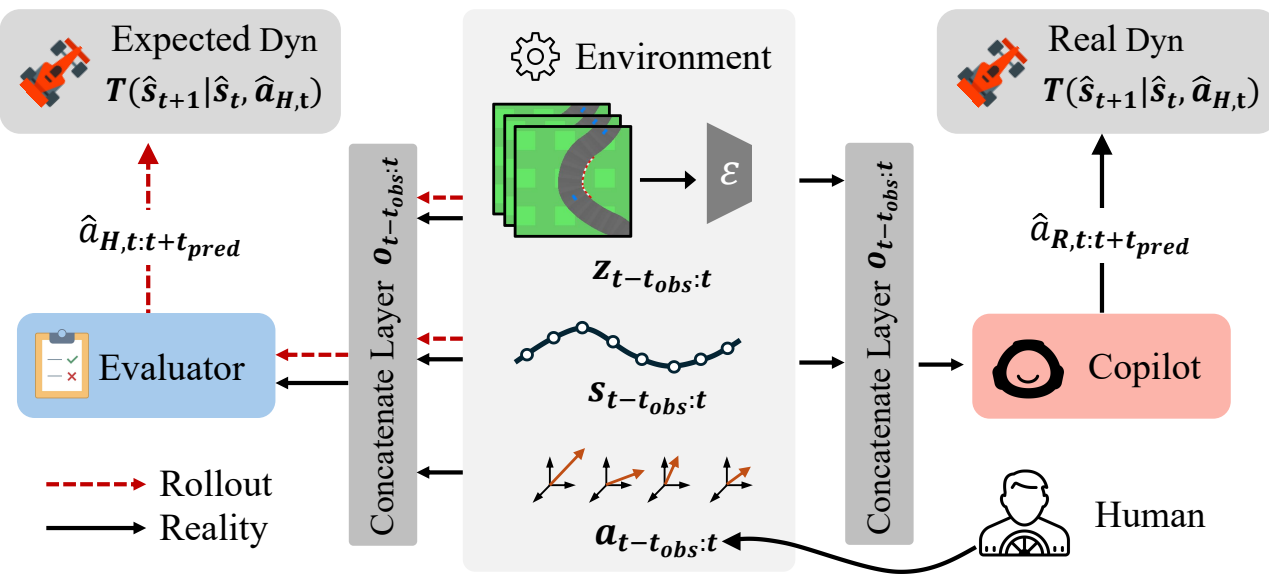

Fig. 2. Closed-loop pipeline of DIFFUSION-SAFE. The evaluator conditions on observation history $\mathbf{O}_t$ and past human actions $\mathbf{A}_t^H$ to predict human-intent sequences, rolled out through the expected dynamics for risk assessment. The copilot conditions only on $\mathbf{O}_t$ to generate safe plans, executed through the real dynamics in closed loop.

models (Fig. 2) that jointly handle risk detection, safe plan generation, and smooth control transfer.

*Contributions.* (i) A closed-loop shared autonomy framework pairing an evaluator for multimodal intent prediction and risk triggering with a safety-guided copilot that enforces safety *during* plan generation via certificate-gradient guidance. (ii) A diffusion-native handover mechanism based on partial diffusion that achieves smooth control transfer without action-level convex mixing. (iii) Validation in simulation and on a real ROS-based race car, demonstrating a 93.0%/87.0% (sim/real) handover success rate with smooth transitions.

## II. METHOD

### A. Problem Formulation and Approach Overview

We consider a discrete-time system $\mathbf{s}_{t+1} = F(\mathbf{s}_t, \mathbf{a}_t)$ with state $\mathbf{s}_t \in S$ and action $\mathbf{a}_t \in A$. DIFFUSION-SAFE uses two diffusion models that represent distributions over short-horizon action sequences.

**Evaluator (multimodal human-intent predictor).** Given an observation-history window $\mathbf{O}_t$ and a past human-action window $\mathbf{A}_t^H$, the evaluator models a multimodal distribution over human-intent plans of horizon $t_{\text{pred}}$:

$$\hat{\mathbf{A}}_t^H \sim p_{\theta_H}\left(\cdot \mid \mathbf{O}_t, \mathbf{A}_t^H\right). \tag{1}$$

Because diffusion models natively represent multimodal distributions, drawing $N$ samples from Eq. 1 yields diverse intent hypotheses without explicit mode enumeration.

**Copilot (expert planner).** The copilot models a distribution over expert plans conditioned on the same observation window:

$$\hat{\mathbf{A}}_t^R \sim p_{\theta_R}(\cdot \mid \mathbf{O}_t)\,. \tag{2}$$

**Safety certificate and probabilistic trigger.** Let $V : \mathbb{R}^2 \to \mathbb{R}$ be a map-based safety certificate (Sec. II-F) with $V(\mathbf{p}) > 0$ inside the drivable surface and $V(\mathbf{p}) \leq 0$ outside; the safe set is $\mathcal{S} = \{\mathbf{p} : V(\mathbf{p}) > 0\}$. For any candidate plan $\mathbf{A}$, let $\tilde{\mathbf{p}}_{t+k}^{\mathbf{A}}$ denote the position reached by rolling out $\mathbf{A}$ through a kinematic surrogate $\hat{F}$ (Sec. II-F), and define the minimum horizon certificate

$$R(\mathbf{s}_t, \mathbf{A}) = \min_{k \in \{1,\dots,t_{\text{pred}}\}} V(\tilde{\mathbf{p}}_{t+k}^{\mathbf{A}}). \tag{3}$$

A planned chunk is safe if $R > 0$. We trigger assistance using a probabilistic risk score

$$\mathcal{R}_t = \mathbb{P}_{\mathbf{A} \sim p_{\theta_H}(\cdot|\mathbf{O}_t, \mathbf{A}_t^H)}\big(R(\mathbf{s}_t, \mathbf{A}) \leq 0\big), \tag{4}$$

and activate assistance when $\mathcal{R}_t > \eta$ for a threshold $\eta \in (0, 1)$.

**Safety-guided copilot.** We augment $p_{\theta_R}$ (Eq. 2) with gradient-based safety guidance, injecting the certificate $V$ directly into the reverse denoising process. We define a trajectory-level safety cost

$$J_{\text{safe}}(\mathbf{A}) = \sum_{i=1}^{t_{\text{pred}}} V(\tilde{\mathbf{p}}_{t+i}^{\mathbf{A}}), \tag{5}$$

whose gradient $\nabla J_{\text{safe}}$ steers each denoising step toward high-$V$ (safe) regions (details in Sec. II-D, II-F). The resulting *safety-guided* copilot distribution is

$$\hat{\mathbf{A}}_t^R \sim p^{\text{guide}}(\cdot \mid \mathbf{O}_t). \tag{6}$$

**Diffusion-native handover via partial diffusion.** When assistance is activated at a replanning instant $t_j$, we realize takeover by partial diffusion: we forward-noise a human-intent plan for $k_\rho$ steps and then run Algorithm 1 from $k_\rho \to 0$, producing a plan whose deviation from the human intent is controlled by $\rho_{t_j} = \frac{k_\rho}{K}$ (Sec. II-E). The resulting distribution is $p^{\text{guide}}(\cdot \mid \mathbf{O}_t;\, \rho_{t_j})$, which reduces to Eq.6 when $\rho_{t_j} = 1$. Since the safety guidance operates at every reverse step, the plan is steered toward safe regions throughout the handover. During takeover, $\rho_{t_j}$ increases gradually across replanning cycles until $\rho_{t_j} = 1$ (full copilot control).

**Execution policy during assistance.** When assistance is inactive the human retains full control; once activated, at each replanning instant $t_j$ the safety-guided copilot generates a fresh plan $\hat{\mathbf{A}}_{t_j}^R \sim p^{\text{guide}}(\cdot \mid \mathbf{O}_{t_j}; \rho_{t_j})$ and the first $t_{\text{pred}}$ actions are executed before the next replan. Because safety is enforced during the reverse process and execution follows the copilot plan without action-level convex mixing, this design mitigates the classical "mixed-unsafe" pitfall under nonconvex safe-action sets (Sec. II-F).

### B. Simulation Description and Setup

We conduct our simulation experiments in *CarRacing-v2* from OpenAI Gym [14], a lightweight 2D racing environment where a car navigates procedurally generated tracks. We choose this environment for its simplicity and interactive nature, while still capturing nonlinear vehicle–terrain interactions via the Box2D physics engine [15]. At each time step ($\Delta t = 0.1$ s), the simulator advances dynamics using an action vector $\mathbf{a}_t = [a_{s,t},\, a_{v,t}]^\top$, where $a_s \in [-1, 1]$ controls steering and $a_v \in [-1, 1]$ is a longitudinal command combining throttle ($a_v > 0$) and braking ($a_v < 0$). User demonstrations are collected via keyboard/joystick input as described in Sec. II-G. Representative simulation tracks are shown in Fig. 3 (top).

### C. Construction of Conditioning Information

We extract a 96×96 RGB patch ahead of the vehicle from the top-down global frame (with the car near the patch bottom) and encode it with a lightweight autoencoder into a latent vector $\mathbf{z}_t \in \mathbb{R}^{128}$, reducing diffusion input dimensionality and inference cost [16]. The observation for both models at time $t$ is $\mathbf{o}_t =$ $[\mathbf{s}_t^\top, \mathbf{z}_t^\top]^\top$, where $\mathbf{s}_t = [\mathbf{p}_t^\top, \theta_t, \mathbf{v}_t^\top]^\top$ with planar position $\mathbf{p}_t \in \mathbb{R}^2$, heading $\theta_t \in \mathbb{R}$, and planar velocity $\mathbf{v}_t \in \mathbb{R}^2$.

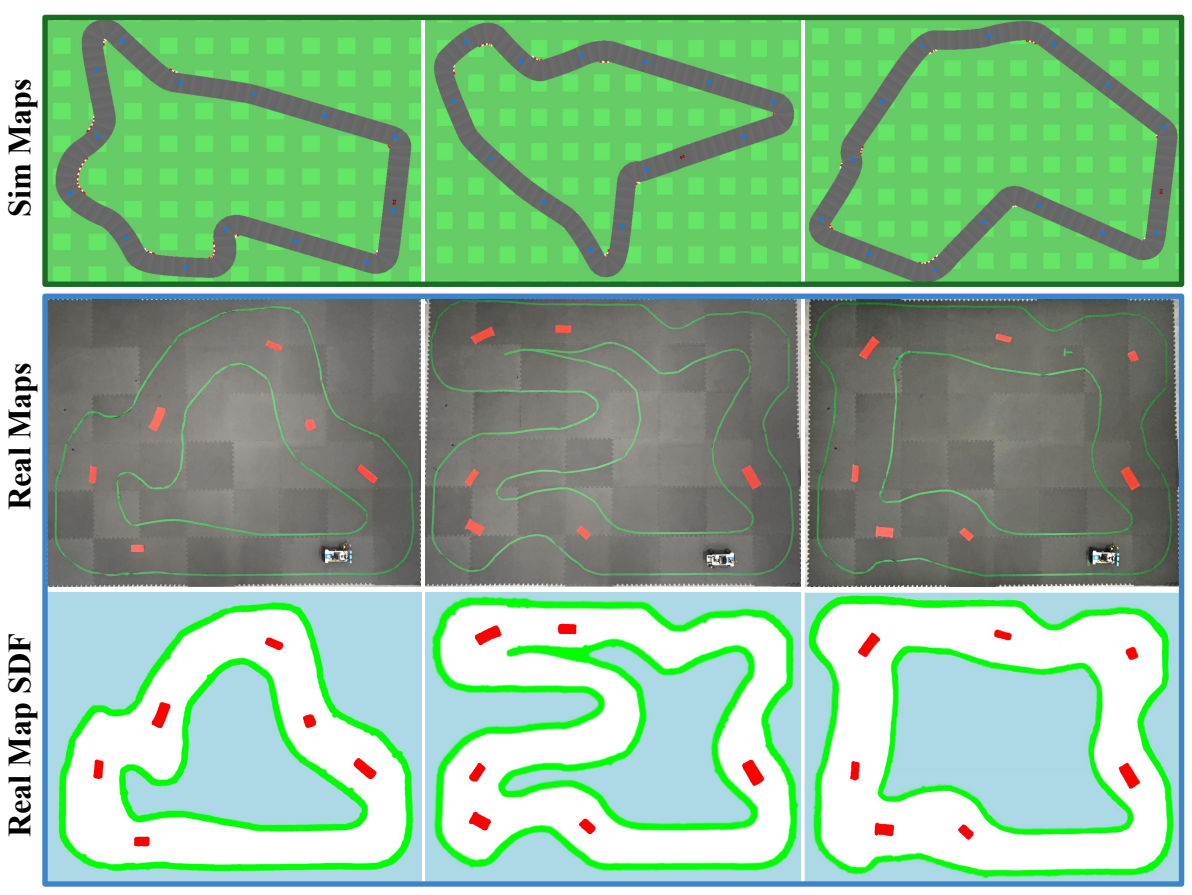

Fig. 3. Track examples and safety certificate visualization. **Top:** randomly generated simulation tracks in CarRacing-v2. **Middle:** real-world tracks built with tape boundaries and obstacles. **Bottom:** signed distance field (SDF) computed from a real track map.

To capture short-term context, we represent observations and actions using fixed-length windows: the stacked observation history $\mathbf{O}_t = [\mathbf{o}_{t-t_{\text{obs}}+1}, \ldots, \mathbf{o}_t]$, the stacked human action history $\mathbf{A}_t^H = [\mathbf{a}_{t-t_{\text{obs}}+1}^H, \ldots, \mathbf{a}_t^H]$, and a predicted action sequence of length $t_{\text{pred}}$: $\hat{\mathbf{A}}_t = [\hat{\mathbf{a}}_t, \ldots, \hat{\mathbf{a}}_{t+t_{\text{pred}}-1}]$. Throughout the paper, lowercase bold symbols (e.g., $\mathbf{o}_t, \mathbf{a}_t$) denote single-step vectors, and uppercase bold symbols (e.g., $\mathbf{O}_t, \mathbf{A}_t^H$) denote window-based sequences.

### D. Diffusion Pipeline for Action Sequence Prediction

In this section, we describe the conditional diffusion model used for action sequence prediction and show how safety guidance is incorporated into the copilot's reverse process.

A diffusion model captures a probability distribution by inverting a forward process that gradually adds Gaussian noise to a data sample $\mathbf{x}^0$ according to a predefined cosine schedule [17]. For the reverse (denoising) process, a neural network $\boldsymbol{\epsilon}_\theta$ takes the noisy input $\mathbf{x}^k$, the diffusion step $k$, and optional conditioning context as input and learns to predict the added noise. To generate a conditional sample, we draw $\mathbf{x}^K \sim \mathcal{N}(\mathbf{0}, \mathbf{I})$ and iteratively denoise for $K$ steps:

$$\mathbf{x}^{k-1} = \alpha_k \big(\mathbf{x}^k - \gamma_k\, \boldsymbol{\epsilon}_\theta(\mathbf{C}, \mathbf{x}^k, k)\big) + \sigma_k\, \mathbf{z}, \quad \mathbf{z} \sim \mathcal{N}(\mathbf{0}, \mathbf{I}), \tag{7}$$

where $\alpha_k, \gamma_k$, and $\sigma_k$ are step-dependent coefficients determined by the noise schedule. We instantiate two conditional models following Eq. 7:

$$\hat{\mathbf{A}}_t^{H,\,k-1} = \alpha_k \big(\hat{\mathbf{A}}_t^{H,k} - \gamma_k\, \boldsymbol{\epsilon}_E(\mathbf{A}_t^H, \mathbf{O}_t, \hat{\mathbf{A}}_t^{H,k}, k)\big) + \sigma_k\, \mathbf{z}, \tag{8}$$

$$\hat{\mathbf{A}}_t^{R,\,k-1} = \alpha_k \big(\hat{\mathbf{A}}_t^{R,k} - \gamma_k\, \boldsymbol{\epsilon}_C(\mathbf{O}_t, \hat{\mathbf{A}}_t^{R,k}, k)\big) + \sigma_k\, \mathbf{z}, \tag{9}$$

where $\boldsymbol{\epsilon}_E$ and $\boldsymbol{\epsilon}_C$ are the denoising networks for the evaluator and copilot, respectively. The evaluator conditions on both historical observations $\mathbf{O}_t$ and human actions $\mathbf{A}_t^H$ (Eq. 8), whereas the copilot conditions only on $\mathbf{O}_t$ (Eq. 9). We implement $\boldsymbol{\epsilon}_E$ and $\boldsymbol{\epsilon}_C$ as 1-D U-Nets with double convolution residual blocks [18], [19]; conditioning is injected via FiLM layers [20] and the diffusion step via sinusoidal positional embeddings [21].

**Safety-guided reverse process.** As introduced in Sec. II-A, we inject the safety certificate $V$ into the copilot's reverse process via gradient-based guidance (Algorithm 1). At each reverse step $k$, we first obtain a one-step estimate of the clean action plan via Tweedie's formula [22]:

$$\hat{\mathbf{A}}_t^{R,0|k} = \frac{1}{\sqrt{\bar{\alpha}_k}} \left( \hat{\mathbf{A}}_t^{R,k} - \sqrt{1 - \bar{\alpha}_k}\, \boldsymbol{\epsilon}_C(\mathbf{O}_t, \hat{\mathbf{A}}_t^{R,k}, k) \right), \tag{10}$$

where $\bar{\alpha}_k = \prod_{i=1}^k \alpha_i$. Because this estimate provides a meaningful clean-plan approximation even at high noise levels, we can evaluate safety at every reverse step: we roll out $\hat{\mathbf{A}}_t^{R,0|k}$ from the current state $\mathbf{s}_t$ through a differentiable kinematic surrogate $\hat{F}$ (a bicycle model fitted to the training data that approximates the simulator dynamics $F$) to obtain rolled-out positions, evaluate $J_{\text{safe}}$ on these positions, and back-propagate through $\hat{F}$ to obtain the action-space gradient $\nabla_{\hat{\mathbf{A}}} J_{\text{safe}}$. This gradient is combined with the standard DDPM posterior mean of the copilot,

$$\boldsymbol{\mu}_C(\hat{\mathbf{A}}_t^{R,k}, k) = \alpha_k \left( \hat{\mathbf{A}}_t^{R,k} - \gamma_k\, \boldsymbol{\epsilon}_C(\mathbf{O}_t, \hat{\mathbf{A}}_t^{R,k}, k) \right), \tag{11}$$

to form the safety-guided reverse step:

$$\hat{\mathbf{A}}_t^{R,k-1} = \boldsymbol{\mu}_C + \lambda(k)\, \nabla_{\hat{\mathbf{A}}} J_{\text{safe}}\left(\hat{\mathbf{A}}_t^{R,0|k}\right) + \sigma_k\, \mathbf{z}, \tag{12}$$

where $\mathbf{z} \sim \mathcal{N}(\mathbf{0}, \mathbf{I})$ and $\lambda(k) = \lambda_0 \cdot \frac{k}{K}$ (we set $\lambda_0 = 0.2$). Guidance is therefore strongest at early (high-noise) steps and vanishes as $k \to 0$, preserving fine plan structure near convergence. Eq.12 replaces Eq.9 whenever the copilot is invoked, including within the partial-diffusion handover chain.

**Algorithm 1:** Safety-Guided Copilot Reverse Process

**Input** : State $\mathbf{s}_t$, observations $\mathbf{O}_t$, human plan $\hat{\mathbf{A}}_t^{H,0}$ (when $k_\rho < K$), SDF grid $V$, start step $k_\rho$, scale $\lambda_0$

**Output** : Denoised action plan $\hat{\mathbf{A}}_t^{R,0}$

1 **if** $k_\rho < K$ **then**
2 $\quad \boldsymbol{\xi} \leftarrow \mathcal{N}(\mathbf{0}, \mathbf{I})$ // Partial diffusion handover
3 $\quad \hat{\mathbf{A}}_t^{R,k_\rho} \leftarrow \sqrt{\bar{\alpha}_{k_\rho}}\, \hat{\mathbf{A}}_t^{H,0} + \sqrt{1 - \bar{\alpha}_{k_\rho}}\, \boldsymbol{\xi}$
4 **else**
5 $\quad \hat{\mathbf{A}}_t^{R,K} \leftarrow \mathcal{N}(\mathbf{0}, \mathbf{I})$ // Full copilot
6 **for** $k \leftarrow k_\rho$ **to** 1 **do**
7 $\quad \hat{\mathbf{A}}_t^{R,0|k} \leftarrow \frac{1}{\sqrt{\bar{\alpha}_k}} \left( \hat{\mathbf{A}}_t^{R,k} - \sqrt{1 - \bar{\alpha}_k}\, \boldsymbol{\epsilon}_C(\mathbf{O}_t, \hat{\mathbf{A}}_t^{R,k}, k) \right)$ // Tweedie estimate
8 $\quad \{\tilde{\mathbf{p}}_{t+i}\}_{i=1}^{t_{\text{pred}}} \leftarrow \text{BicycleRollout}(\mathbf{s}_t, \hat{\mathbf{A}}_t^{R,0|k})$ // Differentiable rollout
9 $\quad J_{\text{safe}} \leftarrow \sum_{i=1}^{t_{\text{pred}}} \text{BilinearSDF}(V, \tilde{\mathbf{p}}_{t+i})$ // Safety cost
10 $\quad \mathbf{g} \leftarrow \nabla_{\hat{\mathbf{A}}_t^{R,0|k}} J_{\text{safe}}$ // Backprop
11 $\quad \boldsymbol{\mu}_C \leftarrow \alpha_k \left( \hat{\mathbf{A}}_t^{R,k} - \gamma_k\, \boldsymbol{\epsilon}_C(\mathbf{O}_t, \hat{\mathbf{A}}_t^{R,k}, k) \right)$ // DDPM posterior mean
12 $\quad \mathbf{z} \leftarrow \mathcal{N}(\mathbf{0}, \mathbf{I})$
13 $\quad \hat{\mathbf{A}}_t^{R,k-1} \leftarrow \boldsymbol{\mu}_C + \lambda_0 \frac{k}{K} \mathbf{g} + \sigma_k \mathbf{z}$ // Guided step
14 **return** $\hat{\mathbf{A}}_t^{R,0}$

## E. Diffusion-native Handover as Partial Diffusion

We view handover as a diffusion-native interpolation between a human-intent plan distribution and the safety-guided copilot distribution. The key idea is to (i) *partially* corrupt a sampled human plan by running a DDPM forward process for $k_\rho$ steps, and then (ii) initialize the *safety-guided* copilot reverse process at that intermediate noise level and denoise back to step 0. This leverages the core mechanism of diffusion models—add noise, then denoise back to the data manifold—to realize a smooth transition from human intent to expert behavior (Fig.1). Because the reverse process projects noisy inputs onto the high-density region of the copilot's learned distribution, the output is a valid expert-like plan; the safety guidance further constrains each step to stay in high-$V$ regions.

We introduce the forward diffusion ratio $\rho = \frac{k_\rho}{K}$, where $k_\rho \in \{0, 1, \ldots, K\}$ is the number of forward (noising) steps and $K$ is the total diffusion horizon. Intuitively, $\rho$ acts as a continuous takeover knob: small $\rho$ keeps the output close to the human plan, while larger $\rho$ makes the reverse chain increasingly prior-driven, pulling the plan toward the copilot's expert mode (Fig. 4(a)).

**Partial forward corruption.** Given a human-intent sample $\hat{\mathbf{A}}_t^{H,0} \in \mathbb{R}^{d_{\text{seq}}}$, we apply $k_\rho$ forward steps using the DDPM closed form:

$$\hat{\mathbf{A}}_t^{H,k_\rho} = \sqrt{\bar{\alpha}_{k_\rho}}\, \hat{\mathbf{A}}_t^{H,0} + \sqrt{1 - \bar{\alpha}_{k_\rho}}\, \boldsymbol{\xi}, \quad \boldsymbol{\xi} \sim \mathcal{N}(\mathbf{0}, \mathbf{I}). \tag{13}$$

**Safety-guided reverse from the noised human plan.** We initialize the copilot reverse chain at $\hat{\mathbf{A}}_t^{R,k_\rho} \leftarrow \hat{\mathbf{A}}_t^{H,k_\rho}$ and run Algorithm 1 from $k_\rho \to 0$ to obtain $\hat{\mathbf{A}}_t^{R,0}$. At each step, the update combines the learned noise prediction $\boldsymbol{\epsilon}_C$ with the safety gradient $\nabla J_{\text{safe}}$, so the denoised plan is simultaneously steered toward the expert prior and kept within safe regions. During takeover, $\rho$ increases linearly from $\rho_{\min}$ to 1. The two endpoints hold by construction: when $\rho = 0$ no noise is injected and the output is the original human plan; when $\rho = 1$ the initialization is pure Gaussian, so the reverse chain samples from the full copilot distribution. For intermediate $\rho$, the noise schedule coefficient $\bar{\alpha}_{k_\rho}$ determines how much human-plan information is preserved: smaller $\bar{\alpha}_{k_\rho}$ (larger $\rho$) retains less, yielding higher conformity to $p^{\text{guide}}$ at the cost of reduced fidelity. We validate this fidelity–conformity tradeoff empirically in Fig.4(b). Since assistance is activated when the evaluator predicts elevated risk, we set a minimum diffusion ratio $\rho_{\min} > 0$ to ensure meaningful correction from the first replanning cycle, and sweep $\rho_{\min}$ to select a Pareto operating point balancing safety and handover smoothness (Fig. 5(b)).

## F. Safety Certificate and Guidance Setup

This section instantiates the safety certificate $V$ used by the safety-guided copilot.

**Map-based safety certificate.** We construct $V$ as a signed distance field (SDF) computed on the known track map, treating both track boundaries and obstacles as impassable regions. For every position $\mathbf{p} = (x, y)$, $V(\mathbf{p})$ returns the signed distance to the nearest boundary of the drivable surface: $V(\mathbf{p}) > 0$ inside and $V(\mathbf{p}) \leq 0$ outside or within an obstacle. The safe set is $\mathcal{S} = \{\mathbf{p} : V(\mathbf{p}) > 0\}$. The bottom row of Fig. 3 visualises the resulting SDF for three real-world tracks, where white denotes the safe interior ($V > 0$), green bands mark the track boundary ($V \approx 0$), light-blue areas are off-road terrain ($V < 0$), and red blocks indicate obstacles embedded in the SDF as non-drivable zones.

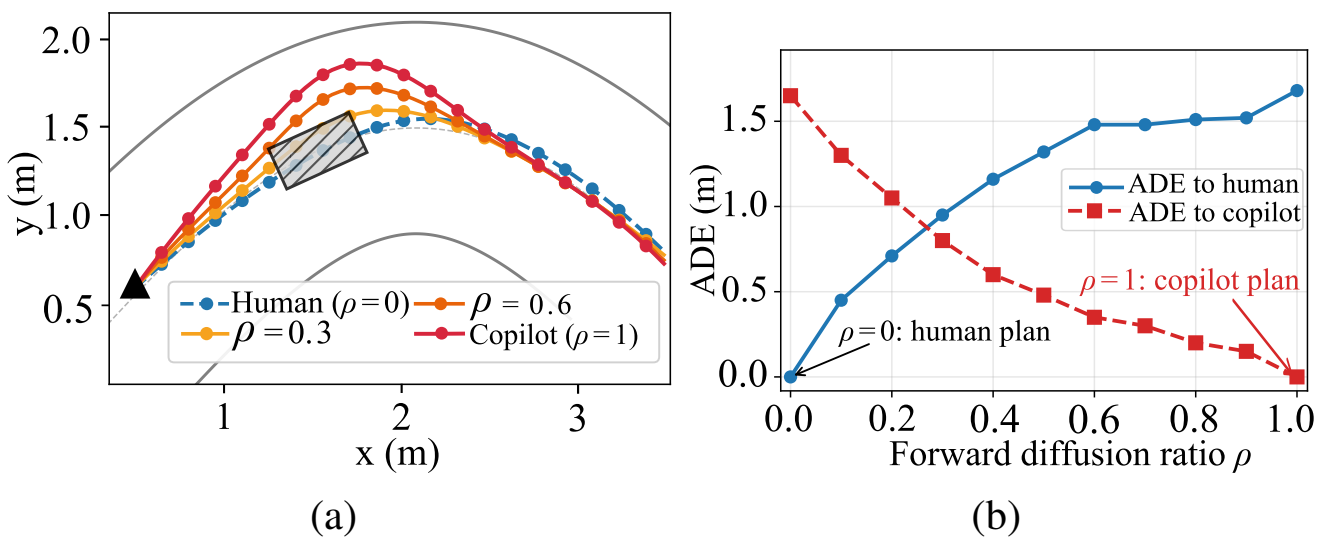

Fig. 4. Effect of the forward-diffusion ratio $\rho$. (a) The human plan ($\rho = 0$) collides with an obstacle; increasing $\rho$ steers the trajectory toward the safe copilot plan ($\rho = 1$), successfully avoiding the obstacle. (b) ADE to the human plan (blue) increases with $\rho$ while ADE to the copilot plan (red) decreases, confirming smooth interpolation between the two endpoints.

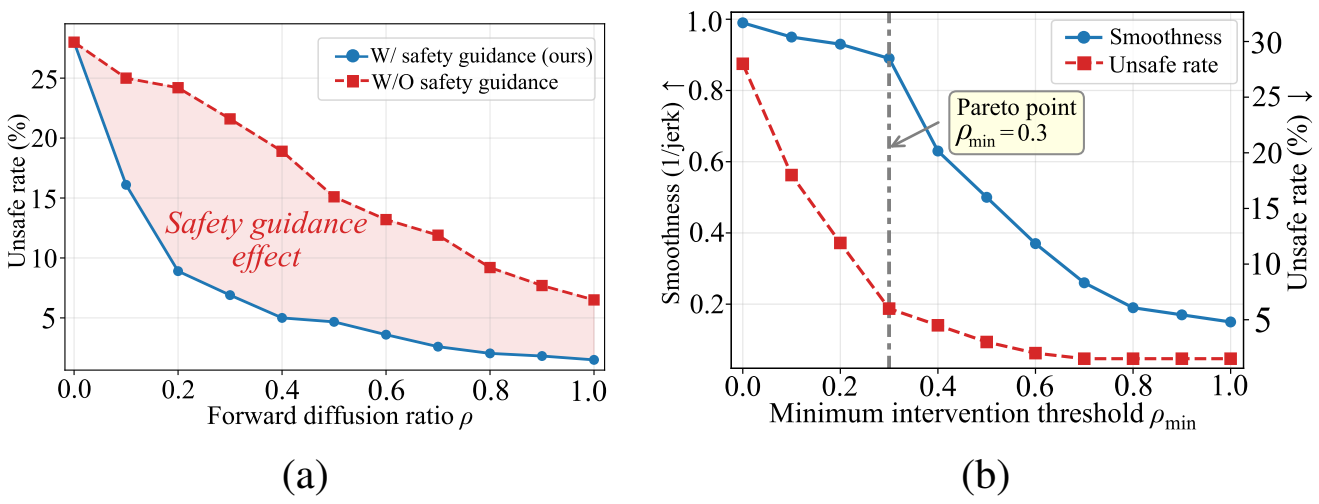

Fig. 5. **Safety and intervention threshold analysis.** (a) Safety violation rate across the full $\rho$ range. With safety guidance (blue), the unsafe rate stays below 3% for all $\rho$; without guidance (red), violations rise sharply as the copilot prior alone cannot cover all safety-critical scenarios. (b) Pareto tradeoff for selecting $\rho_{\min}$. Smoothness (blue) decreases with $\rho_{\min}$, while the unsafe rate (red) saturates beyond $\rho_{\min} = 0.3$. We select $\rho_{\min} = 0.3$ as the operating point.

The trajectory-level safety cost $J_{\text{safe}}$ sums $V$ over the predicted rollout so that its gradient $\nabla_{\hat{\mathbf{A}}} J_{\text{safe}}$ steers the copilot's denoising steps toward high-$V$ (safe) regions. To make this gradient well-defined, we query the SDF grid with bilinear interpolation, ensuring that $V(\mathbf{p})$ and its spatial gradient are continuous and differentiable. Differentiability is likewise required for the rollout itself: because the true dynamics $F$ is non-differentiable and unavailable at deployment, we fit a differentiable kinematic surrogate $\hat{F}$ based on a bicycle model whose parameters are identified from training rollouts, and use $\hat{F}$ for all inference-time rollouts, including both the probabilistic trigger (Eq.4) and the safety-guided reverse process (Algorithm 1). In real-world deployment, the world-frame pose $\mathbf{p}_t^{\text{world}}$ is first projected to pixel coordinates via a homography (Sec. III) before querying the precomputed SDF.

Because safety guidance operates at every reverse step and execution avoids action-level convex blending, the plan remains in safe regions throughout the handover regardless of $\rho_{t_j}$, mitigating the mixed-unsafe pitfall: Fig. 5(a) confirms that the unsafe rate stays below 3% across all $\rho$ when guidance is active (blue), whereas no guidance (red) leads to sharply rising violations.

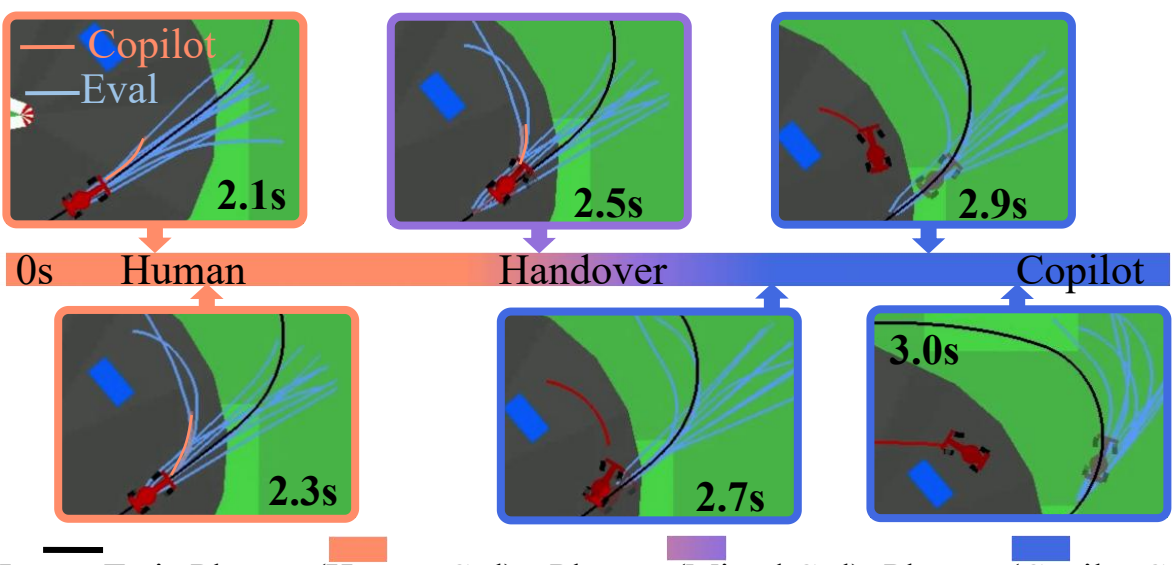

Fig. 6. **Human-to-copilot control transition in *CarRacing-v2*.** Snapshots show a representative handover episode. The human drives normally (0–2.1 s); upon trigger at 2.1 s, copilot authority increases smoothly under mixed control (2.3–2.7 s) until full takeover (≥2.9 s). The shadow car indicates the predicted human trajectory had no intervention occurred.

### G. Dataset and Model Training Details

**Datasets.** Using the *CarRacing-v2* environment, we collect two demonstration datasets via Xbox controllers: an expert dataset $\mathcal{D}_{\text{copilot}}$ and a human dataset $\mathcal{D}_{\text{eval}}$. Each rollout records top-down RGB observations, executed control signals, and simulator state (pose and velocity) in Zarr format. $\mathcal{D}_{\text{eval}}$ contains naturalistic human driving with varying skill levels, including occasional off-road deviations and obstacle collisions. $\mathcal{D}_{\text{copilot}}$ consists of curated demonstrations in which the driver completes the track safely and smoothly; rollouts containing collisions or off-road events are discarded. We use $\mathcal{D}_{\text{copilot}}$ to train the copilot on expert actions, and $\mathcal{D}_{\text{eval}}$ to train the evaluator to model human-intent action sequences.

**Visual encoder.** We train a lightweight autoencoder on the RGB observations using an MSE reconstruction loss. After training, we keep the encoder to map each image to a latent vector $\mathbf{z}_t \in \mathbb{R}^{128}$, which serves as visual conditioning for both diffusion policies.

**Copilot vs. evaluator training.** Both policies are trained to denoise future action sequences from demonstrations (diffusion ELBO objective), but differ in their conditioning. The evaluator learns a human-intent predictor from $\mathcal{D}_{\text{eval}}$, conditioning on both observation history and past human actions to capture the driver's style and intent, including occasional imperfect behaviors. The copilot, in contrast, is trained on $\mathcal{D}_{\text{copilot}}$ to generate expert plans conditioned solely on observations. Crucially, the safety guidance is applied *at inference only*: the gradient $\nabla J_{\text{safe}}$ is injected into the copilot's reverse process without retraining, so a single trained copilot can be paired with different safety certificates.

**Horizons.** We use different observation/prediction horizons for the evaluator and copilot: the evaluator benefits from longer-horizon intent prediction, while the copilot uses a shorter horizon for responsiveness and computational efficiency. We ablate these horizons in Sec. IV-C.

Throughout all experiments we use $K$=50 diffusion steps, draw $N$=15 evaluator samples for risk assessment, and set the trigger threshold $\eta$=0.5; we found consistent performance for $\eta \in [0.4, 0.6]$ and $N \geq 10$. The evaluator uses observation/prediction horizons $t_{\text{obs}} = 20$, $t_{\text{pred}} = 30$; the copilot uses $t_{\text{obs}} = 10$, $t_{\text{pred}} = 15$. We set $\rho_{\min} = 0.3$ (Sec. II-E).

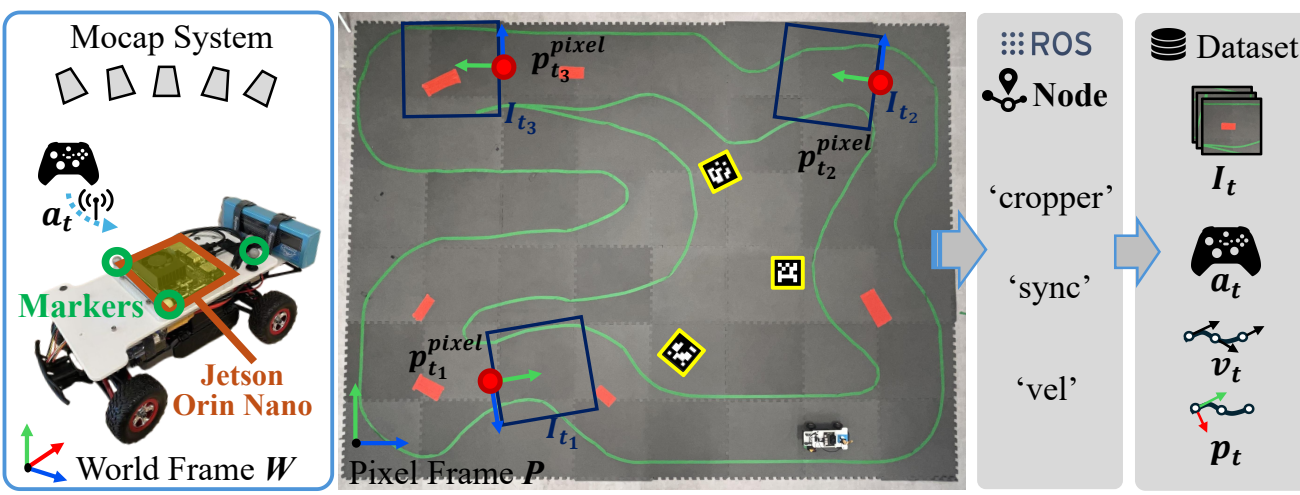

Fig. 7. **Real-world experimental setup. Left:** ROS-controlled race car with Jetson Orin Nano and OptiTrack reflective markers. **Middle:** Overhead view of the track arena; the motion-capture system provides the world-frame pose $\mathbf{p}_t^{\text{world}}$, which is projected to pixel coordinates $\mathbf{p}_t^{\text{pixel}}$ to crop ego-centric observations $I_t$. **Right:** ROS data-collection pipeline; three nodes produce time-aligned tuples $(I_t, \mathbf{a}_t, \mathbf{v}_t, \mathbf{p}_t)$ for training and evaluation.

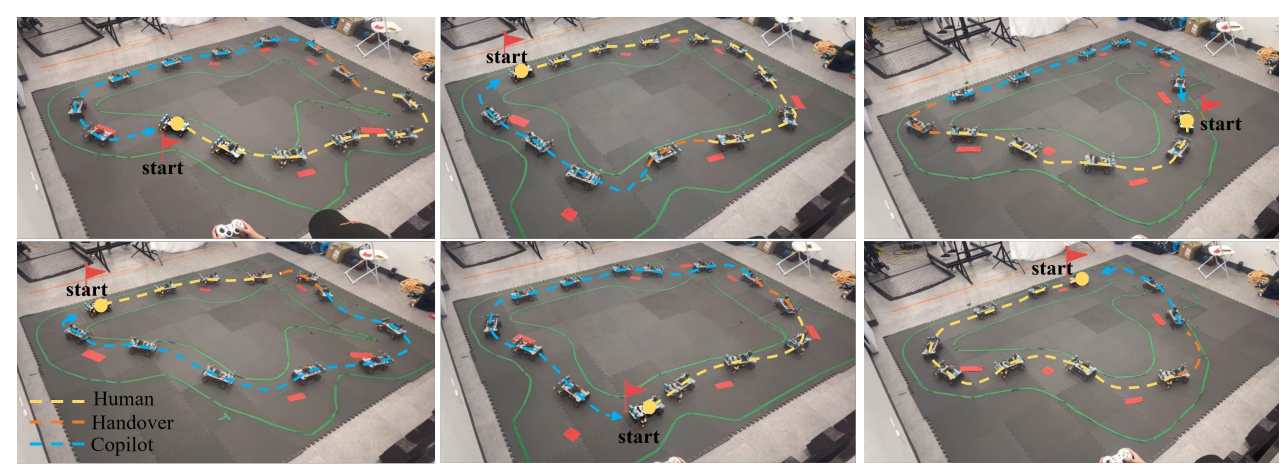

Fig. 8. **Handover demonstrations on unseen maps**. Each column corresponds to a different test track; rows within the same column vary the human driver's starting position and driving strategy, yielding distinct handover scenarios.

## III. Real-World Experiments

### A. Platform and Motion Capture

We evaluate our framework on a ROS-controlled race car with onboard compute (Jetson Orin Nano), tracked by a 13-camera OptiTrack system (Fig. 7). At each time step $t$, the motion-capture pipeline provides the vehicle pose on the ground plane, $\mathbf{p}_t^{\text{world}} = (x_t, y_t, \theta_t)$, streamed to ROS via VRPN. A `cropper` node projects $\mathbf{p}_t^{\text{world}}$ to pixel coordinates and extracts an ego-centric image observation $I_t$ from the overhead camera, while a `vel` node computes finite-difference velocities $\mathbf{v}_t$. A `sync` node time-aligns all streams to produce the training tuple $(I_t, \mathbf{a}_t, \mathbf{v}_t, \mathbf{p}_t)$.

### B. Top-Down Virtual Camera and Patch Cropping

To avoid mounting an onboard camera, we use a single calibrated top-down photograph of the track as a global map. We compute a planar homography $\mathbf{H} \in \mathbb{R}^{3\times3}$ from world-plane coordinates to image pixels using four point correspondences (AprilTag corners in the image and their mocap-measured positions on the ground). For a planar position $(x_t, y_t)$, the pixel location is $[u_t, v_t, 1]^\top \propto \mathbf{H}[x_t, y_t, 1]^\top$. and $(u_t, v_t) = \Pi(\mathbf{H}[x_t, y_t, 1]^\top)$ where $\Pi(\cdot)$ denotes homogeneous normalization. We then crop a $60 \times 60$ patch centered at $(u_t, v_t)$ and rotated by the mocap heading $\theta_t$, yielding a simulation-style top-down observation image $I_t$ in real time (Fig. 3).

### C. Synchronization and Logging

Mocap runs at 100 Hz while joystick commands run at 10 Hz. We timestamp both streams and associate each action with the nearest (or interpolated) mocap pose to form synchronized tuples $(I_t, \mathbf{p}_t^{\text{world}}, \mathbf{a}_t)$. We store demonstrations in the same

Zarr as Sec. II-G, with $\mathbf{a}_t$ (steer/longitudinal command), $I_t$ (cropped patch), and $\mathbf{p}_t^{\text{world}}$ (pose/state). Tracks are built using tape boundaries and obstacles (Fig. 3, middle).

## IV. Evaluation

### A. Evaluation Metrics

We evaluate the evaluator and the copilot using metrics adapted from nuPlan [23]. The evaluator is assessed by human-intent similarity, the copilot by safety-related classification and rollout outcomes, and the full handover pipeline by end-to-end success, safety, and smoothness.

**Human Driving Similarity (Evaluator).** We use minADE-K and minAOE-K ($K$=15) to measure how closely the evaluator's predictions align with human driving behavior. For each test instance, we sample $K$ trajectories from the evaluator and report the minimum average displacement error (minADE-K) and minimum average orientation error (minAOE-K) against the ground-truth human future trajectory. Lower values indicate better coverage of human intent.

**F1 Score (Copilot).** We report the F1 score (harmonic mean of precision and recall) to evaluate the copilot's ability to detect unsafe trajectories. We treat **unsafe off-road** as the positive class and **safe on-road** as the negative class, and compute precision/recall/F1 from the copilot's predicted rollouts. We evaluate on 20 simulated and 20 real tracks, with 10 trials per track.

**Safety (Copilot).** We additionally report rollout-level safety rates: collision rate and off-road rate. Let $N_{\text{traj}}$ be the number of sampled rollouts, $N_{\text{coll}}$ the number of rollouts that collide with obstacles, and $N_{\text{off}}$ the number of rollouts that go off-road. Then

$$Collision = \frac{N_{\text{coll}}}{N_{\text{traj}}}, \qquad Off_Road = \frac{N_{\text{off}}}{N_{\text{traj}}}. \tag{14}$$

Lower values indicate safer copilot predictions.

**Computation Time (Evaluator & Copilot).** Computation time measures inference-time sampling latency (seconds), i.e., the duration to generate a $t_{\text{pred}}$-step trajectory after receiving the input window. We report it for both evaluator and copilot. Because both models predict multi-step action chunks executed in open loop before the next replan, inference occurs only at replanning instants and the measured latencies are well within the respective replanning intervals.

**Handover Success Rate.** We define a handover as successful if the vehicle remains within the safe set $\mathcal{S}$ throughout the entire handover window, i.e., $V(\mathbf{p}_t) > 0$ for all $t$ from the moment assistance is triggered until $\rho_{t_j} = 1$. The success rate is the fraction of successful handovers over all triggered handover episodes.

**Handover Smoothness.** We quantify action-level smoothness during the handover window using the mean absolute jerk:

$$Jerk = \frac{1}{T_h - 1} \sum_{t=t_{\text{start}}}^{t_{\text{end}}-1} \|\mathbf{a}_{t+1} - \mathbf{a}_t\|, \tag{15}$$

where $T_h$ is the number of steps in the handover window. Lower values indicate smoother control transitions.

**Handover Fidelity.** We measure how closely the handover output preserves the original human intent using the negative average displacement error between the executed plan and the human-intent plan over the handover window:

$$Fidelity = -\frac{1}{T_h} \sum_{t=t_{\text{start}}}^{t_{\text{end}}} \|\mathbf{a}_t^R - \mathbf{a}_t^H\|. \tag{16}$$

Higher values (closer to zero) indicate that the copilot plan stays closer to the human intent. This metric complements safety: an ideal handover corrects only what is unsafe while preserving the driver's intended maneuver as much as possible.

TABLE I

Ablation study for evaluator in sim. The last row reports the default configuration in real for reference. **Bold** = Best, *Italic* = Second best.

| Evaluator | Horizon Obs/Pred | minADE-K ↓ | minAOE-K (rad) ↓ | Comput. (s) ↓ |
|---|---|---|---|---|
| Ours (Dflt.) | 20/30 | *0.12* ± 0.02 | *0.13* ± 0.04 | 0.27 ± 0.05 |
| w/o position | 20/30 | 0.29 ± 0.04 | 0.43 ± 0.09 | 0.21 ± 0.04 |
| w/o action | 20/30 | 0.27 ± 0.03 | 0.26 ± 0.03 | 0.20 ± 0.05 |
| w/o vis. | 20/30 | 0.36 ± 0.09 | 0.51 ± 0.07 | **0.13** ± 0.03 |
| Abl. Obs 1 | 15/30 | 0.18 ± 0.02 | 0.15 ± 0.04 | *0.19* ± 0.06 |
| Abl. Obs 2 | 25/30 | 0.13 ± 0.03 | 0.14 ± 0.02 | 0.32 ± 0.04 |
| Abl. Pred 1 | 20/25 | **0.11** ± 0.02 | **0.12** ± 0.04 | 0.26 ± 0.06 |
| Abl. Pred 2 | 20/35 | 0.19 ± 0.04 | 0.19 ± 0.07 | 0.29 ± 0.04 |
| Ours (Real) | 20/30 | 0.18 ± 0.02 | 0.16 ± 0.02 | 0.36 ± 0.03 |

TABLE II

Ablation study for copilot in sim. The last row reports the default configuration in real for reference. **Bold** = Best, *Italic* = Second best.

| Copilot | Horizon Obs/Pred | F1 ↑ | Collision ↓ | Off-Rd. ↓ | Comput. (s) ↓ |
|---|---|---|---|---|---|
| Ours (Dflt.) | 10/15 | *0.97* | **0.02** | **0.01** | 0.18±0.03 |
| w/o position | 10/15 | 0.52 | 0.21 | 0.16 | ***0.15***±0.02 |
| w/o vis. | 10/15 | 0.48 | 0.53 | 0.43 | **0.09**±0.02 |
| Abl. Obs 1 | 5/15 | 0.88 | 0.06 | 0.03 | 0.16±0.03 |
| Abl. Obs 2 | 15/15 | 0.95 | *0.03* | **0.01** | 0.22±0.05 |
| Abl. Pred 1 | 10/10 | **0.98** | **0.02** | **0.01** | 0.16±0.03 |
| Abl. Pred 2 | 10/20 | 0.91 | 0.10 | *0.07* | 0.23±0.02 |
| Ours (Real) | 10/15 | 0.85 | 0.08 | 0.05 | 0.28±0.02 |

### B. Comparison with Baselines

We compare against two multimodal baselines: LSTM-GMM [24] and Behavior Transformers (BET) [25]. Both are trained on the same datasets and horizons as our models, serving as drop-in replacements for the diffusion backbone in either the evaluator or copilot role. Because neither supports the partial forward-noising mechanism of Sec. II-E, we compare them on component-level metrics only (Tables IV, V) and not on the full handover pipeline.

**LSTM-GMM.** LSTM-GMM combines LSTM for temporal feature extraction with GMM to model continuous multimodal action distributions. Notice that we need to pre-define the number of clusters for GMM or k-means steps.

**BET.** BET models employ transformers to model multimodal behaviors by learning from demonstrations. Note that this method also requires the predefined number of modes.

TABLE III
HANDOVER COMPARISON IN SIMULATION AND REAL WORLD. **BOLD** = BEST, *ITALIC* = SECOND BEST.

| Method | Success (%) ↑ | Off-Rd. (%) ↓ | Collision (%) ↓ | Jerk ↓ | Fidelity ↑ |
|---|---|---|---|---|---|
| Ours (full) | **93.0** | **3.0** | **4.0** | **0.08**±0.02 | −**0.14**±0.04 |
| w/o guide | 84.0 | 6.0 | 10.0 | *0.09*±0.03 | −*0.22*±0.05 |
| w/o partial | *89.0* | *5.0* | *6.0* | 0.52±0.10 | −0.87±0.12 |
| SimpleBlend | 76.0 | 8.0 | 16.0 | 0.23±0.06 | −0.25±0.06 |
| Ours (Real) | 87.0 | 5.0 | 8.0 | 0.10±0.03 | −0.31±0.07 |

TABLE IV
BASELINE COMPARISON FOR EVALUATOR IN SIM. **BOLD** = BEST, *ITALIC* = SECOND BEST.

| Evaluator | minADE-K ↓ | minAOE-K (rad) ↓ | Comput. (s) ↓ |
|---|---|---|---|
| Ours | **0.12**±0.02 | **0.13**±0.04 | 0.27±0.05 |
| LSTM-GMM | 0.35±0.03 | *0.35*±0.06 | **0.05**±0.03 |
| BET | 0.39±0.10 | 0.38±0.03 | *0.11*±0.02 |

### C. Ablations and Handover Evaluation

We perform ablation studies on the evaluator and copilot components (Tables I and II) and on the full handover pipeline (Table III) in simulation; the best configuration (*Ours*) is additionally validated on the real platform. Figs. 6 and 8 show representative handover episodes on unseen maps in simulation and real-world settings, respectively.

**Component ablations.** We ablate conditioning inputs and horizons for both models (Tables I, II). For the evaluator, all three inputs contribute: removing vision causes the largest drop (minADE-K triples), followed by position and action history. The horizon 20/30 balances accuracy and latency—shorter observation (15) degrades minADE-K by 50%, while longer prediction (35) hurts accuracy due to compounding error. A shorter prediction horizon (25) achieves marginally lower minADE-K, but we select 30 to extend the risk-assessment window. For the copilot, removing position or vision drops F1 below 0.55. A prediction horizon of 10 matches the default in F1 but offers fewer steps for safety guidance; extending to 20 degrades F1 to 0.91. We select 10/15, as the longer horizon provides more correction steps while maintaining low collision and off-road rates.

**End-to-end handover comparison.** Table III evaluates the full handover pipeline under four configurations: (i) *Ours*: safety-guided copilot with partial diffusion; (ii) *w/o guidance*: partial diffusion with the standard copilot ($\lambda(k) = 0$); (iii) *w/o partial diffusion (Hard switch)*: copilot assumes full control instantaneously ($\rho$ jumps from 0 to 1); (iv) *Simple blending*: action-level linear interpolation $\mathbf{a}_{\text{blend}} = k\,\mathbf{a}^H + (1-k)\,\mathbf{a}^C$, where $k$ decreases over time. Our full method achieves the highest success rate and lowest off-road and collision rates. Removing safety guidance degrades safety, confirming that the copilot prior alone does not sufficiently cover all safety-critical scenarios. Hard switching approaches our method in safety (89.0% vs. 93.0% success) but produces significantly higher jerk (6.5×) due to the abrupt transition. Simple blending yields the worst safety outcome—action-level interpolation under nonconvex safe-action sets can steer the vehicle off-track even when both individual actions are safe (Figs. 9, 10).

TABLE V
BASELINE COMPARISON FOR COPILOT IN SIM. **BOLD** = BEST, *ITALIC* = SECOND BEST.

| Copilot | F1 ↑ | Collision ↓ | Off-Rd ↓ | Comput. (s) ↓ |
|---|---|---|---|---|
| Ours | **0.97** | **0.02** | **0.01** | 0.18 ± 0.03 |
| LSTM-GMM | 0.65 | *0.11* | *0.12* | **0.11** ± 0.02 |
| BET | 0.74 | 0.13 | 0.14 | *0.13* ± 0.03 |

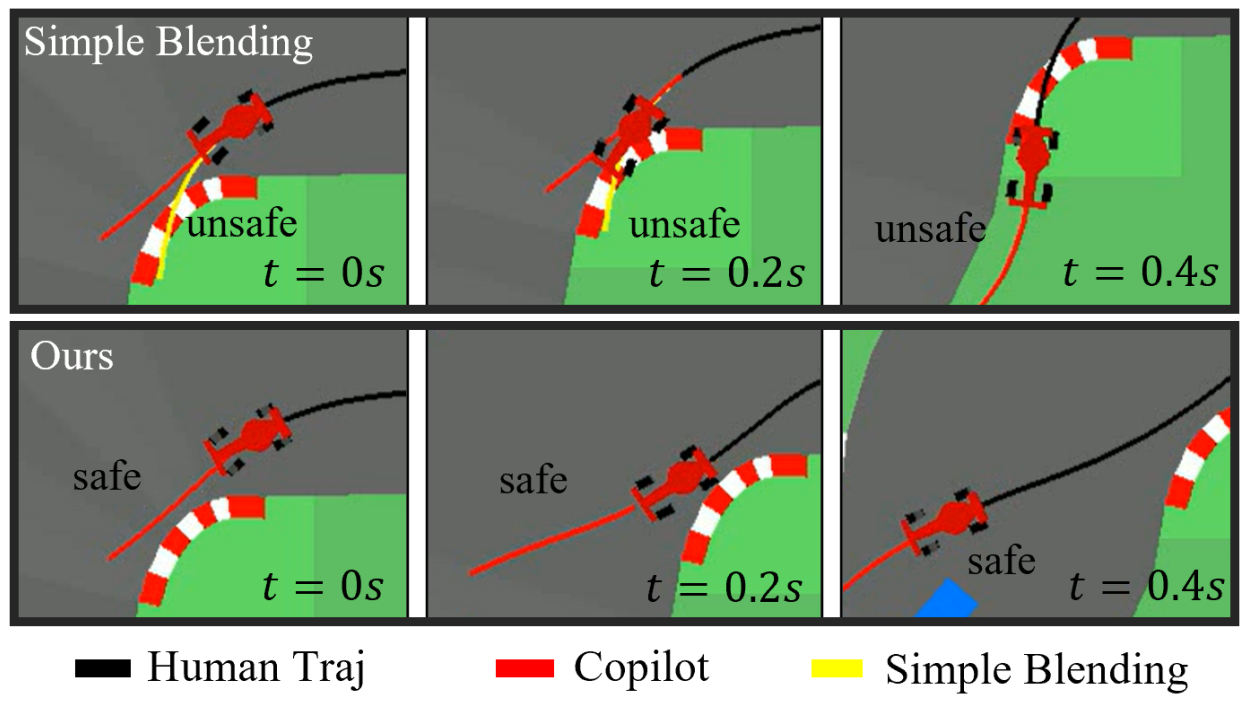

Fig. 9. Top-down trajectory snapshots comparing simple action blending (top row) and our partial diffusion handover (bottom row) under identical initial states and human inputs at three time steps.

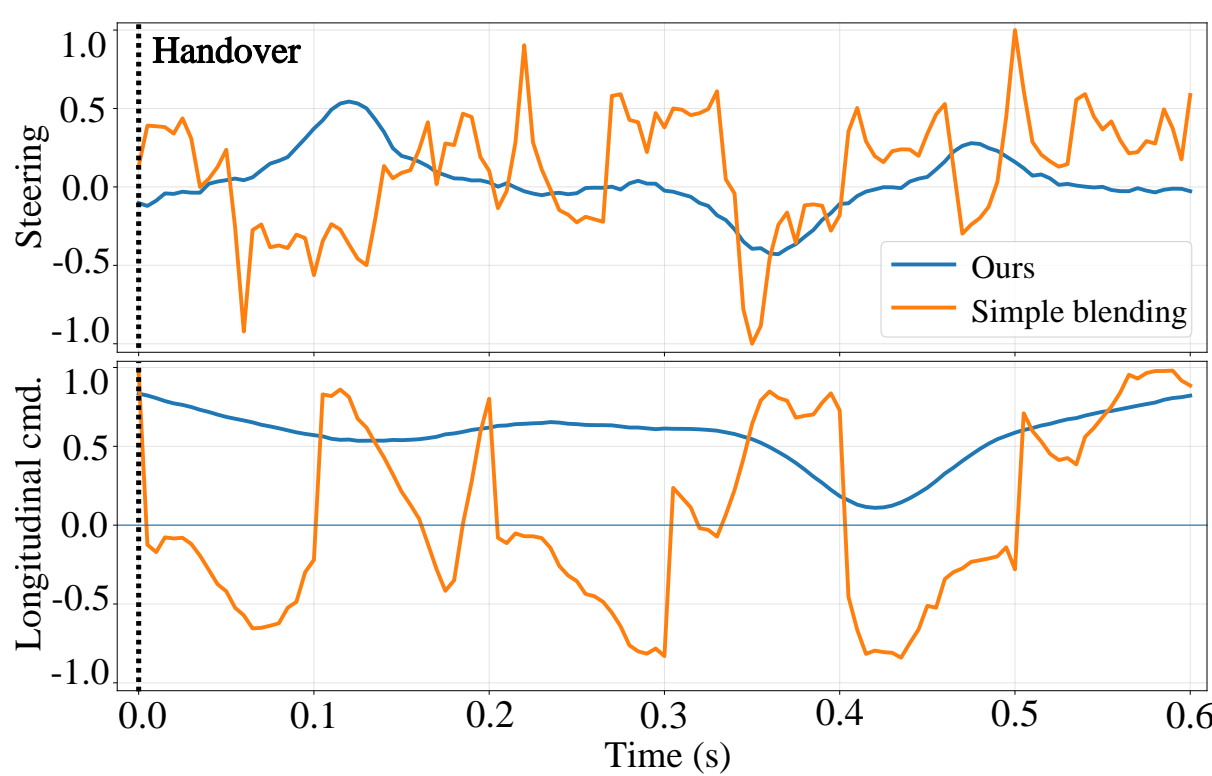

Fig. 10. **Action-level smoothness during handover.** We compare our diffusion-native handover (blue curves) against simple action blending (orange curves). Plots show steering $a_s$ and longitudinal commands $a_v$ over the handover window (trigger → takeover completion).

### D. Failure Case Analysis

Despite the high overall success rate, we observe residual failures in both simulation (7.0%) and real-world (13.0%) deployments. We categorize them below.

**Simulation.** The dominant failure mode is *late triggering*: on tracks with abrupt, high-curvature turns, the evaluator's risk score $\mathcal{R}_t$ exceeds the threshold $\eta$ only after the vehicle has already entered a state from which the copilot's prediction horizon $t_{\text{pred}}$ is too short to recover. A secondary mode is *guidance saturation*: when the human plan is far off-track and $\rho$ is still small (i.e., $k_\rho \ll K$), the cumulative safety correction over the few available reverse steps is too small to overcome the large initial deviation, particularly near narrow track sections where the feasible corridor is thin.

**Real world.** Real-world failures are additionally driven by sensing and actuation artifacts. Motion-capture noise and the 10 Hz joystick rate introduce 1–2 step state-estimation lag (0.1–0.2 s), compounding the late-triggering mode.

## V. Conclusion

We presented Diffusion-SAFE, a shared autonomy framework that unifies probabilistic risk detection, safety-guided plan generation, and smooth control transfer within a single diffusion-based pipeline. Experiments show that combining safety guidance with partial diffusion achieves a 93.0% handover success rate in simulation and 87.0% on the real platform while maintaining low jerk, outperforming both hard switching and action-level blending; ablations confirm that neither component alone suffices. A key insight is that operating in plan space via partial diffusion sidesteps the mixed-unsafe failure mode inherent to action-level blending under nonconvex safe-action sets, while the continuous $\rho$ knob provides a principled alternative to binary hard switching.

The current framework assumes structured environments with known maps and 2D top-down observations. Future work will address these limitations along three axes: (i) extending the safety certificate to 3D perception pipelines (e.g., LiDAR or multi-camera BEV representations), (ii) learning the certificate from data rather than precomputing it from a map, enabling deployment in unknown environments, and (iii) investigating adaptive $\rho$ scheduling that adjusts the takeover aggressiveness based on real-time risk severity rather than a fixed ramp.

## References

[1] S. Reddy, A. D. Dragan, and S. Levine, "Shared autonomy via deep reinforcement learning," *arXiv preprint arXiv:1802.01744*, 2018.

[2] L. Xiong, C. B. Chng, C. K. Chui, P. Yu, and Y. Li, "Shared control of a medical robot with haptic guidance," *International Journal of Computer Assisted Radiology and Surgery*, vol. 12, pp. 137–147, 2017.

[3] K. Vellenga, H. J. Steinhauer, A. Karlsson, G. Falkman, A. Rhodin, and A. C. Koppisetty, "Driver intention recognition: State-of-the-art review," *IEEE Open Journal of Intelligent Transportation Systems*, vol. 3, pp. 602–616, 2022.

[4] M. Flad, J. Otten, S. Schwab, and S. Hohmann, "Steering driver assistance system: A systematic cooperative shared control design approach," in *2014 IEEE International Conference on Systems, Man, and Cybernetics (SMC)*. IEEE, 2014, pp. 3585–3592.

[5] P. Trautman, "Assistive planning in complex, dynamic environments: a probabilistic approach," *arXiv preprint arXiv:1506.06784*, 2015.

[6] E. Shi, T. M. Gasser, A. Seeck, and R. Auerswald, "The principles of operation framework: A comprehensive classification concept for automated driving functions," *SAE International Journal of Connected and Automated Vehicles*, vol. 3, no. 12-03-01-0003, pp. 27–37, 2020.

[7] M. Marcano, S. Díaz, J. Pérez, and E. Irigoyen, "A review of shared control for automated vehicles: Theory and applications," *IEEE Transactions on Human-Machine Systems*, vol. 50, no. 6, pp. 475–491, 2020.

[8] C. Meng, Y. He, Y. Song, J. Song, J. Wu, J.-Y. Zhu, and S. Ermon, "SDEdit: Guided image synthesis and editing with stochastic differential equations," in *ICLR*, 2022.

[9] T. Yoneda, L. Sun, G. Yang, B. Stadie, and M. Walter, "To the noise and back: Diffusion for shared autonomy," in *Robotics: Science and Systems (RSS)*, 2023.

[10] C. Chi, S. Feng, Y. Du, Z. Xu, E. Cousineau, B. Burchfiel, and S. Song, "Diffusion policy: Visuomotor policy learning via action diffusion," in *Robotics: Science and Systems (RSS)*, 2023.

[11] M. Janner, Y. Du, J. B. Tenenbaum, and S. Levine, "Planning with diffusion for flexible behavior synthesis," in *International Conference on Machine Learning (ICML)*, 2022.

[12] W. Xiao, T.-H. Wang, C. Gan, and D. Rus, "Safediffuser: Safe planning with diffusion probabilistic models," *arXiv preprint arXiv:2306.00148*, 2023.

[13] J. Zhang, L. Zhao, A. Papachristodoulou, and J. Umenberger, "Constrained diffusers for safe planning and control," *arXiv preprint arXiv:2506.12544*, 2025.

[14] G. Brockman, V. Cheung, L. Petterski, J. Schneider, J. Schulman, J. Tang, and W. Zaremba, "Openai gym," *arXiv preprint arXiv:1606.01540*, 2016.

[15] E. Catto, "box2d," https://github.com/erincatto/box2d, 2023, accessed: 2023.

[16] R. Rombach, A. Blattmann, D. Lorenz, P. Esser, and B. Ommer, "High-resolution image synthesis with latent diffusion models," *Proceedings of the IEEE/CVF Conference on Computer Vision and Pattern Recognition (CVPR)*, pp. 10 684–10 695, 2022.

[17] A. Q. Nichol and P. Dhariwal, "Improved denoising diffusion probabilistic models," in *International Conference on Machine Learning (ICML)*, 2021.

[18] J. Ho, A. Jain, and P. Abbeel, "Denoising diffusion probabilistic models," *Advances in neural information processing systems*, vol. 33, pp. 6840–6851, 2020.

[19] S. Zagoruyko and N. Komodakis, "Wide residual networks," *arXiv preprint arXiv:1605.07146*, 2016.

[20] E. Perez, F. Strub, H. D. Vries, V. Dumoulin, and A. Courville, "Film: Visual reasoning with a general conditioning layer," *Proceedings of the AAAI Conference on Artificial Intelligence*, vol. 32, no. 1, 2018.

[21] A. Vaswani, N. Shazeer, N. Parmar, J. Uszkoreit, L. Jones, A. N. Gomez, L. Kaiser, and I. Polosukhin, "Attention is all you need," *Advances in Neural Information Processing Systems*, vol. 30, 2017.

[22] B. Efron, "Tweedie's formula and selection bias," *Journal of the American Statistical Association*, vol. 106, no. 496, pp. 1602–1614, 2011.

[23] H. Caesar, J. Kabzan, K. S. Tan, W. K. Fong, E. Wolff, A. Lang, L. Fletcher, O. Beijbom, and S. Omari, "nuPlan: A closed-loop ML-based planning benchmark for autonomous vehicles," in *NeurIPS Datasets and Benchmarks Track*, 2021.

[24] A. Mandlekar, D. Xu, J. Wong, S. Nasiriany, C. Wang, R. Kulkarni, L. Fei-Fei, S. Savarese, Y. Zhu, and R. Martín-Martín, "What matters in learning from offline human demonstrations for robot manipulation," *IEEE Robotics and Automation Letters*, vol. 6, no. 4, pp. 6522–6529, 2021.

[25] N. M. Shafiullah, Z. Cui, A. A. Altanzaya, and L. Pinto, "Behavior transformers: Cloning $k$ modes with one stone," *IEEE Robotics and Automation Letters*, vol. 7, no. 4, pp. 10 456–10 463, 2022.